%% file: main.tex
\documentclass{article}

\newif\ifarxivversion
\arxivversiontrue

\usepackage{iclr2027_conference,times}

\input{preamble}

\definecolor{iclrblue}{rgb}{0.13,0.32,0.55}
\usepackage[breaklinks,colorlinks,allcolors=iclrblue]{hyperref}
\usepackage{url}

\title{CSWAM: Better Causal Semantic Representations for Out-of-Distribution Generalization in World Action Models}
\author{%
  \makebox[0.96\textwidth][c]{\resizebox{0.96\textwidth}{!}{\textbf{%
    Tianbin Liu$^{1}$, Jian Zhu$^{1,*,\dagger}$, Taiyi Su$^{1}$,
    Jianjun Zhang$^{1,2}$, Chong Ma$^{1,2}$, Zitai Huang$^{1,2}$, Weiyi Lu$^{1}$,
    Yi Xu$^{1,\dagger}$}}}\\[4pt]
  \makebox[0.96\textwidth][c]{\normalfont
    $^{1}$AIRC, Midea Group \quad $^{2}$Tongji University}\\
  \makebox[0.96\textwidth][c]{\normalfont\small\texttt{%
    liutb27@midea.com, jianzhu823@gmail.com, xuyi42@midea.com}}\\[2pt]
  \makebox[0.96\textwidth][c]{\normalfont\small
    Code: \href{https://github.com/midea-ai/CSWAM}{\texttt{github.com/midea-ai/CSWAM}}}
}

\iclrfinalcopy

\begin{document}

\maketitle
\fancyhead{}
\renewcommand{\headrulewidth}{0pt}
\begingroup
  \renewcommand{\thefootnote}{\fnsymbol{footnote}}
  \footnotetext[1]{Project lead.}
  \footnotetext[2]{Corresponding author.}
\endgroup

\input{sec/0_abstract}
\input{sec/1_intro}
\input{sec/2_related_work}
\input{sec/3_method}
\input{sec/4_experiments}
\input{sec/6_conclusion}

\bibliographystyle{iclr2027_conference}
\bibliography{main}

\end{document}

%% file: preamble.tex
\usepackage{amsmath,amssymb}
\usepackage{array}
\usepackage{booktabs}
\usepackage{float}
\usepackage{graphicx}
\usepackage{makecell}
\usepackage{microtype}
\usepackage{multirow}
\usepackage{wrapfig}
\usepackage{xcolor}

\newcommand{\method}{CSWAM}

%% file: sec/0_abstract.tex
\begin{abstract}
FastWAM-style world action models enable efficient action-only inference, but
generalize poorly under visual distribution shifts. Their
reconstruction-oriented representations emphasize appearance-specific details,
limiting generalization to unseen scenes and objects. Without observation
history, the model also lacks temporal evidence for robustly identifying
task-relevant state changes and motion in unfamiliar visual conditions. To address these
limitations, we present the
\emph{Causal Semantic World Action Model} (\method), which augments FastWAM
with a causal semantic expert built on V-JEPA~2.1. V-JEPA provides temporally
grounded representations of semantic state changes and motion with less
dependence on appearance-specific details. The expert learns their future
evolution from a sparse history of current and past observations and shares the
history-derived context with both the video and action streams through causal
attention.
At inference, \method{} conditions action denoising on the current
video state and observed semantic history, retaining efficient action-only
inference. We conduct simulation and real-robot experiments to evaluate
generalization under distribution shifts. With embodied pretraining,
\method{} raises Randomized success on RoboTwin~2.0 Clean-to-Randomized transfer
from 10.16\% to 45.18\%, a gain of 35.02 percentage points over FastWAM. Across
two real-robot tasks and three OOD difficulty levels, \method{} improves average
success over FastWAM by 42.5 percentage points, from 27.5\% to 70.0\%.
\end{abstract}

%% file: sec/1_intro.tex
\section{Introduction}
\label{sec:intro}

By jointly modeling future visual states and robot actions, world action models
(WAMs)~\cite{gr2,lingbotva,dreamzero,fastwam,mecowam,dswam,openwam} have emerged as a
promising alternative to vision-language-action (VLA)
models~\cite{rt2,openvla,pi0,demavla} for robot manipulation.
However, WAMs that first predict future video and then generate actions incur
substantial latency at inference. Consequently, a growing line of work adopts a
FastWAM-style paradigm that jointly trains video and action prediction but
generates only actions at deployment~\cite{uva,fastwam,linjepawam}. This design
retains the benefits of video co-training while avoiding iterative future-video
generation at deployment, making the paradigm attractive for robot
control~\cite{fastwam}.

FastWAM-style models perform well when the visual distribution at deployment
resembles that of the training data. However, we find that FastWAM's success rate falls
sharply under visual distribution shifts, such as changes in backgrounds and
object appearances. Two limitations may contribute to this gap: the visual
representation used for action prediction and the absence of observation
history. First, the video stream operates on
reconstruction-oriented VAE latents, which preserve appearance details but are
not explicitly aligned with task-relevant semantics. Changes in background or
object appearance can therefore alter the visual context even when the
underlying manipulation state is unchanged. Second, the action expert receives
only the current observation and cannot use history that reveals motion, state
transitions, and task progress. Without stable semantics and temporal evidence,
the policy can rely on correlations specific to the training scenes that fail
under visual shifts.

Generalization to out-of-distribution (OOD) visual conditions is essential for
real-world deployment, since collecting demonstrations across every background
and object appearance is infeasible. Recent work has introduced pretrained
semantic encoders such as DINO~\cite{dinov3} to make robot policies and WAMs less
sensitive to appearance changes~\cite{robustvisrep,stwam}. These studies highlight the
value of semantic features, but leave open how temporally predictive
representations and observation history should be incorporated into efficient
action-only WAMs. DINO provides strong spatial semantics, but when applied as a
framewise encoder, it does not explicitly represent how a scene evolves. We
therefore use V-JEPA~2.1~\cite{vjepa,vjepa2,vjepa21}, whose video-native
joint-embedding predictive representations more directly capture semantic
state changes and motion while reducing reliance on appearance-specific
details. Paired with causal observation history,
these representations provide both task-relevant semantic cues and richer
temporal evidence for action prediction.

Motivated by this analysis, we introduce \method{}, a Causal Semantic World
Action Model that extends the FastWAM-style video--action architecture with a
history-conditioned semantic expert (Fig.~\ref{fig:overview}). A frozen
V-JEPA~2.1 encoder first extracts semantic features
from a sparse history of current and past observations. The semantic expert
predicts future semantic features to model how task-relevant states evolve over
time. A structured attention mask allows both the video and action streams to
use the observed semantic history while preventing the action stream from
accessing future targets. At deployment, the future target streams are omitted:
the current video state and semantic history are computed once, and only the
action stream is denoised. By sharing the same history-derived semantic and
motion context with both streams, \method{} supports future-video learning and
action prediction while preserving FastWAM's efficient action-only inference.
Experiments in simulation and on real robots demonstrate consistent gains
under visual distribution shifts. Under the standard RoboTwin~2.0 mixed-training
protocol, \method{} achieves 94.7\% success on Clean and 94.3\% on Randomized,
respectively. Under the stricter Clean-to-Randomized transfer setting with
embodied pretraining, \method{} raises OOD Randomized success from 10.16\% to
45.18\%, a gain of 35.02 percentage points over FastWAM. On LIBERO-Plus,
\method{} achieves 77.1\% overall success, exceeding ST-WAM~\cite{stwam} by
4.3 percentage points. Across the two real-robot tasks and three OOD difficulty levels,
\method{} raises average success from 27.5\% to 70.0\%, a gain of 42.5
percentage points over FastWAM.

Our contributions are threefold:
\begin{itemize}
  \item We investigate how temporally predictive representations and
  observation history support OOD generalization in efficient action-only WAMs,
  and design controlled visual-shift evaluations in simulation and on real
  robots.

  \item We propose \method{}, whose V-JEPA~2.1-based causal semantic expert
  predicts future semantic features from observation history and shares the
  history-derived context with both the video and action streams, while
  retaining action-only inference.

  \item Extensive simulation and real-robot experiments show that \method{}
  improves both standard manipulation performance and OOD generalization over
  FastWAM.
\end{itemize}

%% file: sec/2_related_work.tex
\section{Related work}
\label{sec:related}

\subsection{World action models}

World action models (WAMs) learn robot policies jointly with predictive visual
dynamics, using future-state modeling to provide supervision about scene
evolution and action consequences. Representative formulations couple video
and action through joint prediction~\cite{gr2}, shared representations with
decoupled prediction heads~\cite{uva}, or causal autoregressive
diffusion~\cite{lingbotva}. This paradigm has also been scaled with pretrained
video diffusion backbones and heterogeneous robot data~\cite{dreamzero}. More
recent work explores complementary directions, including 4D geometric
priors~\cite{mecowam}, dual-system architectures~\cite{dswam}, and modular
pretraining frameworks~\cite{openwam}.
An important deployment question is whether visual prediction remains active
at inference. The Unified Video Action model allows actions to be decoded
without generating video~\cite{uva}. FastWAM develops this idea into an
action-only WAM: video and action prediction are jointly trained, but
future-video denoising is removed at deployment~\cite{fastwam}. Using a
Mixture-of-Transformers architecture~\cite{mot}, it conditions action denoising
on a fixed context derived from the current observation. Subsequent methods
explore alternative trade-offs between future conditioning and inference
efficiency. LingBot-VA~2.0 continues to predict future latents during
asynchronous closed-loop execution~\cite{lingbotva2}, whereas Faster-WAM
computes compact future representations once and reuses them during action
denoising~\cite{fasterwam}. \method{} instead builds directly on FastWAM's
action-only formulation, expanding its observed context with temporally
predictive V-JEPA history while retaining inference without future prediction.

\begin{figure}[t]
  \centering
  \includegraphics[width=0.94\textwidth]{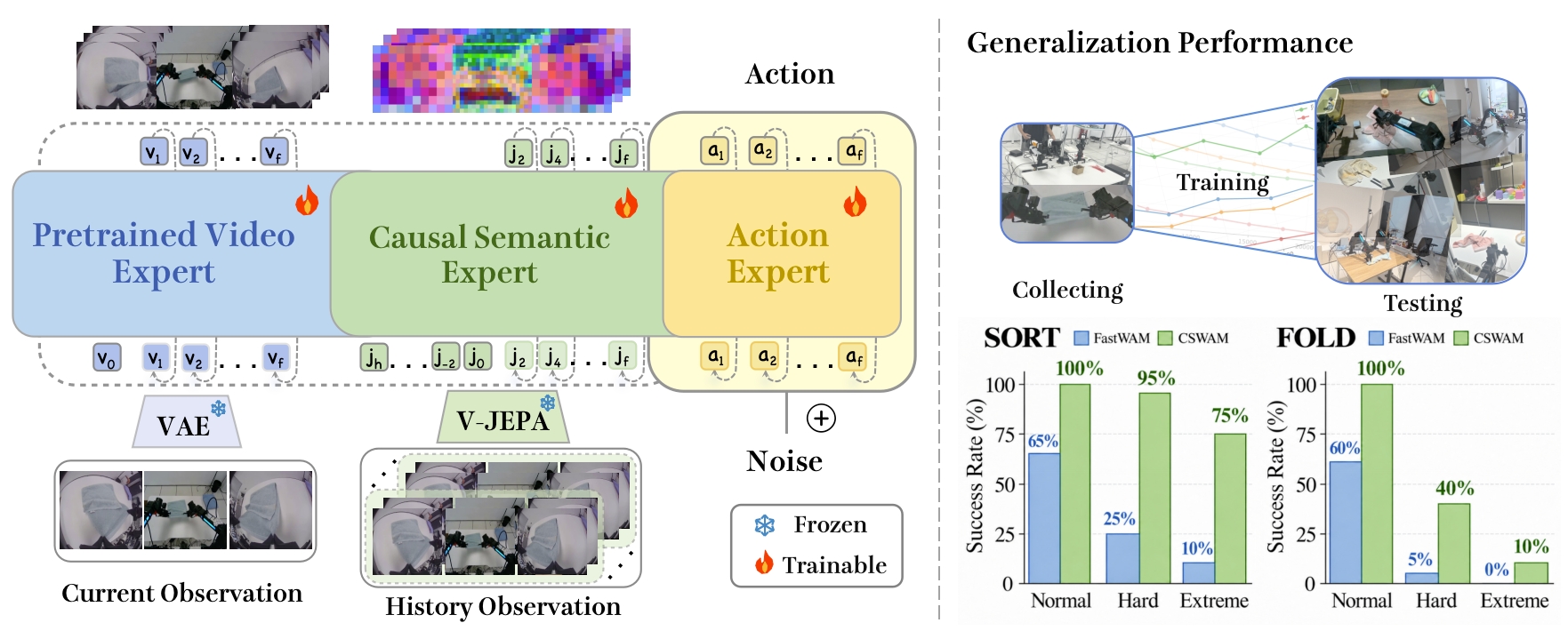}
  \caption{Overview of \method{}. Left: the causal semantic expert encodes
  V-JEPA observation history and shares temporal context with the video and
  action experts. Right: real-robot success under compounded visual shifts.}
  \label{fig:overview}
\end{figure}

\subsection{Visual representations for robot manipulation}

Visual representations determine what information world modeling contributes
to action prediction. Generative WAMs commonly operate in VAE latent spaces
optimized for visual reconstruction and prediction~\cite{dreamzero,fastwam}.
These latents preserve the appearance details required for video generation,
but may also retain nuisance variations unrelated to manipulation. Studies of
robust visuomotor learning show that pretrained semantic representations can
reduce this sensitivity~\cite{robustvisrep}. DINOv3~\cite{dinov3} learns strong
dense visual features through image self-supervision and has been adopted as
semantic supervision in WAMs~\cite{stwam}. When applied as a framewise encoder,
however, it does not explicitly model temporal relations, leaving state changes
and motion to be inferred by the downstream model.
Joint-embedding predictive architectures (JEPAs) learn representations by
predicting target features from visible context rather than reconstructing
pixels. I-JEPA~\cite{ijepa} establishes this principle for images, and
V-JEPA~\cite{vjepa} extended it to video. V-JEPA~2~\cite{vjepa2} scales
action-free video pretraining and demonstrates strong motion understandings, 
while V-JEPA~2.1~\cite{vjepa21} improves spatial grounding through dense context
loss and deep self-supervision. Concurrent JEPA-WAM~\cite{linjepawam} predicts
structured V-JEPA targets using a predictor shared with action generation.
\method{} instead uses a dedicated semantic expert to predict future V-JEPA
features from sparse observation history and shares the history-derived context
with both the video and action streams.

\subsection{OOD generalization in robot manipulation}

Visual distribution shifts remain a persistent challenge for robot policies
trained from demonstrations. Studies of visuomotor behavior cloning show that
even pretrained visual encoders can be sensitive to subtle changes in lighting,
scene texture, and distractor objects, and that in-distribution accuracy alone
does not predict robustness~\cite{robustvisrep}. Scaling to VLAs does not remove
this failure mode; similar failures have been reported under controlled visual
distribution shifts~\cite{liberoplus}. Cross-paradigm evaluations further
suggest that WAMs can benefit from spatiotemporal video priors, but their
robustness depends on how those priors are incorporated into action
prediction~\cite{wamrobustness}.
Within WAMs, recent methods improve robustness through different forms of
future or historical conditioning. ST-WAM~\cite{stwam} combines DINOv3 future
supervision with current-anchored retrieval from semantic history.
Faster-WAM~\cite{fasterwam} retains compact future representations during action
denoising to balance robustness and inference efficiency. \method{} instead
preserves FastWAM's action-only inference while using video-native V-JEPA
history to provide temporally predictive semantic context to both the video and
action streams. Our evaluation separates mixed-training performance from OOD
transfer through Clean-to-Randomized simulation and compounded real-robot
visual shifts.

%% file: sec/3_method.tex
\section{Causal Semantic World Action Model}
\label{sec:method}

\subsection{Overview}

Our goal is to strengthen the observation representation available to a
FastWAM-style policy without giving up its action-only deployment
paradigm~\cite{fastwam}. \method{} retains the video and action streams and adds
a semantic stream derived from observation history. The video stream preserves
generative visual dynamics, while the semantic stream provides temporally
grounded state and motion context that is less tied to the appearance of the
current frame. Separate transformer experts process the three streams and
exchange information through a Mixture-of-Transformers (MoT)
architecture~\cite{mot}, as illustrated in Fig.~\ref{fig:overview}. During
training, future visual and semantic prediction shape these representations
alongside action generation; at deployment, the action expert uses only
information computed from current and past observations.

We formalize the resulting action policy as follows. Let $o_t$ be the current
multi-view observation, $\mathcal{H}_t$ an observation history ending at $o_t$,
and $\ell$ a language instruction. The video encoder
represents the current observation as $z_V^c=E_V(o_t)$, while the frozen
semantic encoder represents the history as $z_J^h=E_J(\mathcal{H}_t)$. Given
these observed features, the deployed policy predicts an action chunk
$a_{t:t+T_a-1}\in\mathbb{R}^{T_a\times d_a}$ according to
\begin{equation}
  \hat a_{t:t+T_a-1}
  \sim p_\theta\!\left(\,\cdot\mid z_V^c,z_J^h,\ell\right),
  \label{eq:policy}
\end{equation}
where $T_a$ is the action horizon and $d_a$ is the action dimension.

\subsection{Causal semantic expert}

The generative video stream must retain fine appearance details to predict
future frames, yet these details can change substantially across environments
without altering the underlying task state. We therefore introduce the causal
semantic expert as a complementary pathway that supplies action prediction
with temporally predictive visual context. Its design combines a frozen
V-JEPA feature space, which emphasizes semantic state changes and motion, with
a sparse causal history that reveals how these states evolve up to the current
observation. The expert learns to extrapolate future semantic features from
this observed history.

To expose temporal changes over a longer interval without increasing $K$, we
sample the history at a stride of $s$ raw video frames,
\begin{equation}
  \mathcal{H}_t
  = \bigl(o_{t-(K-1)s},\ldots,o_{t-s},o_t\bigr).
  \label{eq:history}
\end{equation}
We pass the entire history to a frozen V-JEPA~2.1 encoder~\cite{vjepa21}. Its
input embedding partitions the sampled observations into non-overlapping
temporal tubelets of size $q$, after which all spatiotemporal tokens are
processed jointly. The encoded history is
\begin{equation}
  z_J^h = E_J(\mathcal{H}_t) \in
  \mathbb{R}^{B\times d_J\times T_h\times H_J\times W_J},
  \label{eq:jepa-encoding}
\end{equation}
where $B$ is the batch size, $d_J$ is the encoder feature dimension, and
$T_h\times H_J\times W_J$ is its spatiotemporal feature grid, with
$T_h=K/q$. Each temporal slice therefore aggregates a short local transition,
while video attention integrates evidence across the full observed history
rather than treating its frames as independent descriptors.

The V-JEPA feature grid is spatially dense. Before it enters the semantic
expert, a learnable spatiotemporal projection $P_J$ reduces its spatial token
count, preserves its temporal order, and maps it to the expert's hidden space.
This produces the observed semantic tokens
$J_h=P_J(z_J^h)$.

\ifarxivversion
\else
  \newpage
\fi
\begin{figure}[!t]
  \centering
  \includegraphics[width=0.90\textwidth]{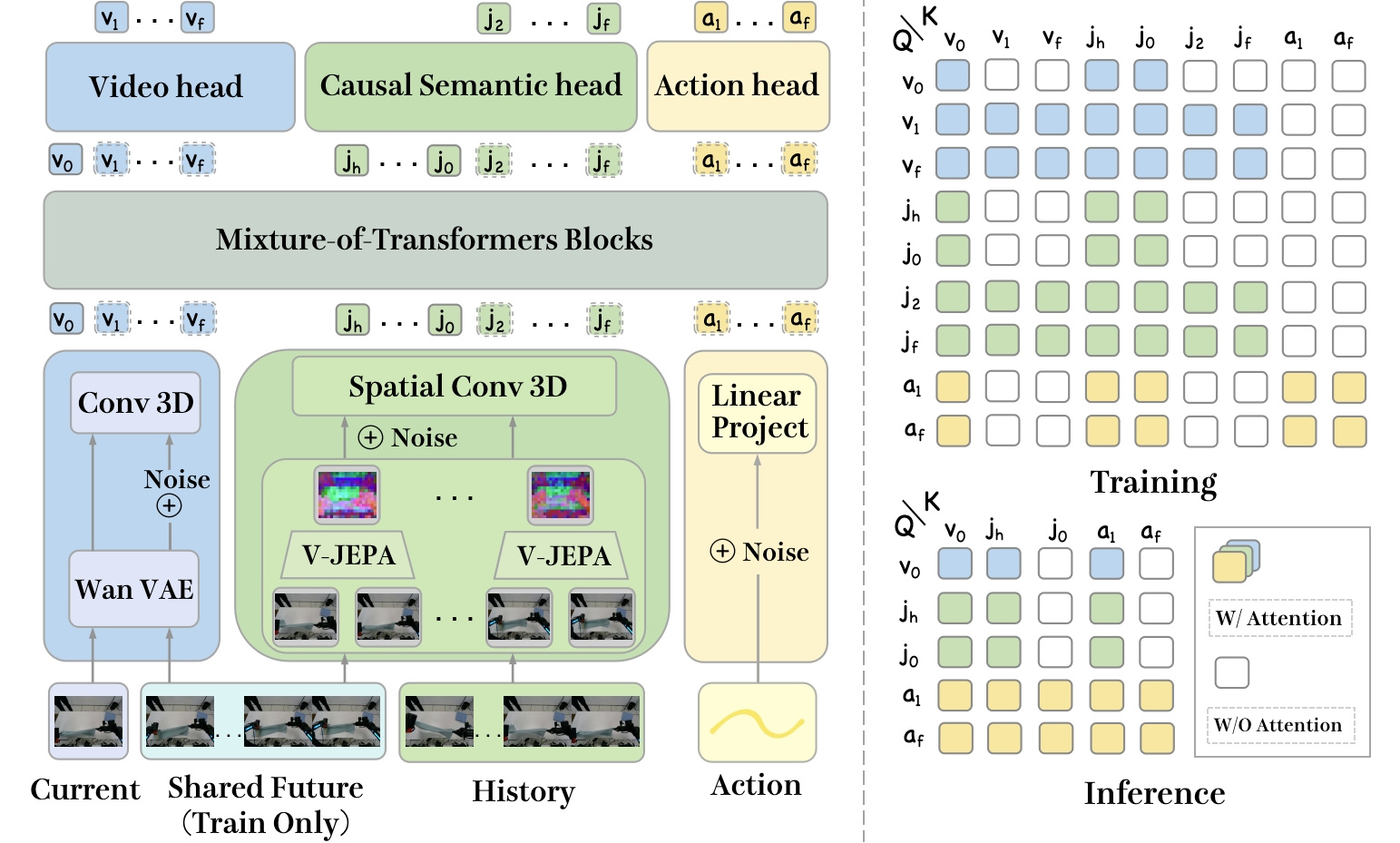}
  \caption{Architecture and attention pattern of \method{}. During training,
  future video and semantic tokens provide auxiliary prediction targets, while
  semantic history conditions both the video and action streams. At inference,
  the model retains observed context and denoises only actions.}
  \label{fig:architecture}
  \vspace{-0.4\baselineskip}
\end{figure}

Encoding history makes temporal evidence available, but does not by itself
require the trainable semantic pathway to capture predictive state evolution.
We therefore train it to extrapolate future V-JEPA features. The same frozen
encoder maps a future clip $\mathcal{F}_t$ to the prediction target
$z_J^f=E_J(\mathcal{F}_t)$, which we perturb at a sampled semantic noise level
$\tau_J$,
\begin{equation}
  z_{J,\tau_J}^f
  = \alpha_{\tau_J}z_J^f+\sigma_{\tau_J}\epsilon_J,
  \qquad \epsilon_J\sim\mathcal{N}(0,I),
  \label{eq:semantic-noise}
\end{equation}
where $\alpha_{\tau_J}$ and $\sigma_{\tau_J}$ define the noise path, and
$J_f=P_J(z_{J,\tau_J}^f)$ denotes the projected future tokens.

We formulate this prediction in the same flow-matching framework as the video
and action streams. A diffusion transformer~\cite{dit} processes $[J_h;J_f]$
and predicts the flow of the future semantic features conditioned on the clean
history. An output head maps its prediction to the target feature space, and
the loss is applied only to the future tokens. This objective encourages the
expert to model semantic state evolution rather than use V-JEPA features as
static conditioning.

\subsection{Cross-stream context integration}

The semantic pathway should inform the original WAM computation rather than
serve only as an auxiliary prediction task. We therefore integrate it with the
video and action pathways through masked MoT attention. The video expert takes
the current VAE latent $z_V^c$ as a clean prefix and learns a flow field for
noised future video latents, retaining the appearance detail and generative
dynamics of the pretrained video model. ActionDiT operates on a noised action
chunk and predicts the corresponding action flow. Let $V_c$, $V_f$, and $A$
denote the current-video, future-video, and action tokens.

Together with the causal semantic expert, these pathways form a three-expert
MoT. Each expert retains its modality-specific projections and feedforward
layers, while masked attention allows information to be exchanged across
streams without collapsing their modality-specific representations. The mask
is designed both to share observed context and to isolate future prediction
targets. In particular, the video tokens attend to the observed V-JEPA history
$J_h$. The semantic history therefore not only conditions the semantic expert,
but also enriches the video representation used to learn future visual
dynamics. During training, the future video and semantic streams can further
exchange information, allowing their complementary prediction objectives to
be learned jointly.

The action expert attends to both the current-video tokens $V_c$ and the
semantic-history tokens $J_h$, combining visual dynamics with history-derived
semantic state and motion. All three experts are also conditioned on language;
Fig.~\ref{fig:architecture} summarizes the resulting information flow.

\subsection{Joint training and action-only inference}

Joint training couples the three representations while retaining a separate
objective for each stream. The video and semantic experts predict future
latent flows, while the action expert predicts the action flow. For stream
$m\in\{V,J,A\}$, let $u_m$ and $\hat u_m$ denote its target and predicted
flow. The joint objective is
\begin{equation}
  \mathcal{L}
  = \sum_{m\in\{V,J,A\}}
    \lambda_m\,
    \mathbb{E}\!\left[w_m(\tau_m)\|\hat u_m-u_m\|_2^2\right],
  \label{eq:objective}
\end{equation}
where $\lambda_m$ balances the three streams, $\tau_m$ is the sampled noise
level, and $w_m(\tau_m)$ is its noise-dependent weight. Each squared error is
averaged over valid elements after applying the padding mask of its modality.
The video and semantic streams share a sampled noise level,
$\tau_V=\tau_J$, aligning the corruption levels of their future targets,
whereas the action stream samples $\tau_A$ independently. Through the shared
MoT attention, the three losses jointly shape the interacting representations;
the V-JEPA encoder remains frozen.

At inference, we discard $V_f$ and $J_f$, compute $V_c$ and $J_h$ once, and
reuse them while the action expert performs iterative denoising.

%% file: sec/4_experiments.tex
\section{Experiments}
\label{sec:experiments}

\subsection{Evaluation protocols and setup}

\paragraph{RoboTwin 2.0.}
We evaluate 50 bimanual tasks from RoboTwin~2.0~\cite{robotwin2} under two
protocols. In
\textbf{RoboTwin2.0-Full}, following the standard multi-task setting used by
Motus, LingBot-VA, FastWAM, and ST-WAM~\cite{motus,lingbotva,fastwam,stwam}, a
single policy is trained on 2,500 Clean and 25,000 Randomized demonstrations
and evaluated in both settings; this measures mixed-training performance rather
than transfer. In \textbf{RoboTwin2.0-Clean2Random}, policies are trained only
on the 2,500 Clean demonstrations and tested on Randomized scenes without
adaptation. We fine-tune for five epochs, select the checkpoint with the lowest
validation loss, and report success over 100 rollouts per task.

Clean2Random is evaluated with and without embodied pretraining (PT). For
Embodied PT, FastWAM is first pretrained on a 5,000-hour robotics corpus, and
\method{} inherits its video- and action-expert weights. The causal semantic
transformer is warm-started from shape-compatible ActionDiT parameters, while
its input and output layers are initialized separately.

\paragraph{LIBERO-Plus.}
We additionally use LIBERO-Plus~\cite{liberoplus}, which evaluates seven
controlled shifts in camera viewpoint, robot initial state, language
instruction, lighting, background texture, sensor noise, and object layout.
Following the benchmark protocol, we report category-wise and overall success
rates. \method{} uses the embodied-pretrained initialization described above.

\begin{figure}[!t]
  \centering
  \includegraphics[width=\textwidth]{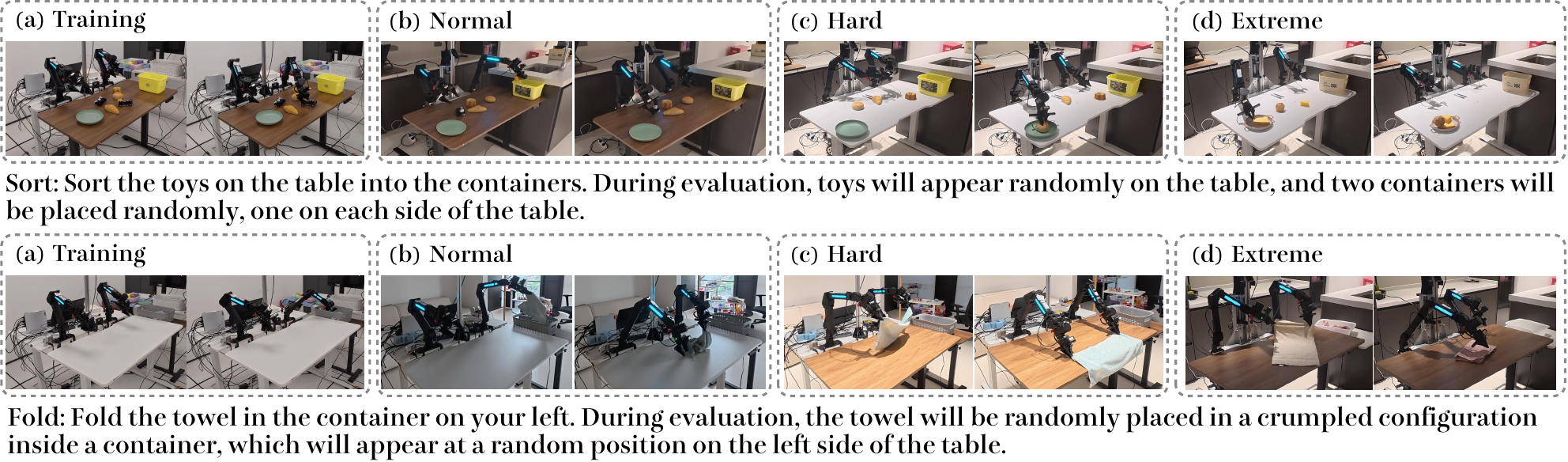}
  \caption{Real-robot evaluation scenes for Sort (top) and Fold (bottom), shown
  for training and the Normal, Hard, and Extreme OOD settings.}
  \label{fig:real-scenes}
  \vspace{-1.5\baselineskip}
\end{figure}

\paragraph{Real robot.}
We use a dual-arm platform with two ARX Robotics manipulators for two
long-horizon tasks: sorting toys into containers (Sort) and folding a towel
(Fold). Both models use the embodied-pretrained initialization above and are
fine-tuned on the same task data---12 hours for Sort and 4 hours for Fold.
We evaluate the shifts in Table~\ref{tab:real-protocol}; ID and OOD are defined
relative to each task's fine-tuning distribution, and object initialization is
randomized at every level. Each task, level, and model is tested for 20 trials
(240 total); we report success rate and mean completion time over successful
trials. Figure~\ref{fig:real-scenes} shows the corresponding scenes.

\ifarxivversion
\else
  \newpage
\fi

\begin{table}[H]
  \vspace{-1.8\baselineskip}
  \begin{minipage}[t]{0.48\textwidth}
    \vspace{0pt}
    \centering
    \caption{Real-robot evaluation settings.}
    \label{tab:real-protocol}
    \small
    \setlength{\tabcolsep}{1.2pt}
    \begin{tabular}{@{}lccccc@{}}
      \toprule
      Level & Env. & Lighting & Tabletop & Object & Container \\
      \midrule
      Normal  & OOD & ID  & ID  & ID  & ID  \\
      Hard    & OOD & OOD & OOD & ID  & ID  \\
      Extreme & OOD & OOD & OOD & OOD & OOD \\
      \bottomrule
    \end{tabular}
  \end{minipage}\hfill
  \begin{minipage}[t]{0.48\textwidth}
    \vspace{0pt}
    \centering
    \addtocounter{table}{1}
    \caption{OOD success (\%) on RoboTwin2.0-Clean2Random. Embodied PT precedes
    task fine-tuning.}
    \label{tab:clean-to-random}
    \small
    \setlength{\tabcolsep}{1.2pt}
    \begin{tabular}{@{}lcccc@{}}
      \toprule
      Method & \makecell{Embodied\\PT} & Clean & Randomized & Average \\
      \midrule
      FastWAM & No  & 63.68 & 1.82  & 32.75 \\
      FastWAM & Yes & 76.68 & 10.16 & 43.42 \\
      \method{} & No  & 78.48 & 9.36  & 43.92 \\
      \method{} & Yes & \textbf{84.46} & \textbf{45.18} & \textbf{64.82} \\
      \bottomrule
    \end{tabular}
    \addtocounter{table}{-2}
  \end{minipage}
  \vspace{-0.4\baselineskip}
\end{table}

\subsection{Simulation: standard performance and OOD transfer}
\label{sec:exp-simulation}

\paragraph{Standard mixed-training performance.}
On RoboTwin2.0-Full, \method{} achieves 94.7\% on Clean, 94.3\% on Randomized,
and 94.5\% on average, the best results among the listed VLA and WAM baselines
(Table~\ref{tab:robotwin}). It improves over FastWAM by 2.6 points on average
and over ST-WAM by 1.6, 1.8, and 1.7 points on Clean, Randomized, and average,
respectively. Because both settings appear in training, these results measure
mixed-training robustness rather than OOD transfer.

\paragraph{Clean-to-Randomized OOD transfer.}
\label{sec:exp-clean-to-random}

Mixed-training performance does not establish transfer to unseen visual
distributions, so we evaluate the stricter RoboTwin2.0-Clean2Random protocol:
both methods are trained only on Clean demonstrations and tested on Randomized
scenes without adaptation (Table~\ref{tab:clean-to-random}). Without embodied
PT, \method{} achieves 43.92\% average success versus 32.75\% for FastWAM, an
11.17-point gain independent of embodied-data pretraining. Embodied PT improves
both models, but its benefit is substantially larger for \method{}: average
success increases by 10.67 points for FastWAM and by 20.90 points for
\method{}, including a 35.82-point gain on Randomized. With embodied PT,
\method{} ultimately outperforms FastWAM by 7.78 points on Clean and 35.02
points on Randomized, raising the average from 43.42\% to 64.82\%. The much
larger improvement on unseen Randomized scenes indicates that embodied priors
and causal semantic modeling are complementary, rather than merely improving
fit to the fine-tuning distribution.

\paragraph{LIBERO-Plus robustness.}
Table~\ref{tab:libero-plus} compares \method{} with VLA and WAM baselines under
the seven LIBERO-Plus shifts. \method{} achieves the highest overall success
rate of 77.1\%, outperforming the strongest baseline, ST-WAM, by 4.3 percentage
points. It performs particularly well under camera-viewpoint and lighting
shifts, reaching 85.9\% and 93.4\%, respectively, while its 52.1\% success under
robot-initial-state shifts identifies variation in the initial robot state as
the most challenging category.

\begin{table}[t]
  \centering
  \caption{Success rate (\%) on RoboTwin2.0-Full under standard mixed Clean and
  Randomized training. Baselines are grouped into VLA and WAM methods.}
  \label{tab:robotwin}
  \footnotesize
  \setlength{\tabcolsep}{3.6pt}
  \begin{tabular}{@{}lccc@{}}
    \toprule
    Method & Clean & Randomized & Average \\
    \midrule
    \multicolumn{4}{c}{\textbf{VLA}} \\
    \midrule
    X-VLA~\cite{xvla} & 72.8 & 72.8 & 72.8 \\
    $\pi_{0.5}$~\cite{pi05} & 82.7 & 76.8 & 79.8 \\
    ABot-M0~\cite{abotm0} & 86.1 & 85.1 & 85.6 \\
    Qwen-VLA~\cite{qwenvla} & 86.1 & 87.2 & 86.7 \\
    Galaxea G0.5~\cite{g05} & 93.7 & 92.8 & 93.3 \\
    Qwen-RobotManip~\cite{qwenrobotmanip} & 93.7 & 94.0 & 93.9 \\
    \midrule
    \multicolumn{4}{c}{\textbf{WAM}} \\
    \midrule
    Motus~\cite{motus} & 88.7 & 87.0 & 87.8 \\
    FastWAM~\cite{fastwam} & 91.9 & 91.8 & 91.9 \\
    LingBot-VA~\cite{lingbotva} & 92.9 & 91.6 & 92.2 \\
    ST-WAM~\cite{stwam} & 93.1 & 92.5 & 92.8 \\
    LingBot-VA 2.0~\cite{lingbotva2} & 93.8 & 93.4 & 93.6 \\
    ABot-M0.5~\cite{abotm05} & 94.0 & 94.2 & 94.1 \\
    \midrule
    \method{} (ours) & \textbf{94.7} & \textbf{94.3} & \textbf{94.5} \\
    \bottomrule
  \end{tabular}
\end{table}

\begin{table}[t]
  \vspace{-1.2\baselineskip}
  \addtocounter{table}{1}
  \centering
  \caption{Success rate (\%) on LIBERO-Plus.}
  \label{tab:libero-plus}
  \footnotesize
  \setlength{\tabcolsep}{1.2pt}
  \begin{tabular}{@{}lccccccccc@{}}
    \toprule
    Method & \makecell{Emb.\\PT} & Camera & Robot & Lang. & Light & BG & Noise & Layout & Overall \\
    \midrule
    UniVLA & Yes & 1.8 & 46.2 & 69.6 & 69.0 & 81.0 & 21.2 & 31.9 & 42.9 \\
    $\pi_0$~\cite{pi0} & Yes & 13.8 & 6.0 & 58.8 & 85.0 & 81.4 & 79.0 & 68.9 & 53.6 \\
    $\pi_0$-FAST & Yes & 65.1 & 21.6 & 61.0 & 73.2 & 73.2 & 74.4 & 68.8 & 61.6 \\
    RIPT-VLA (OFT) & Yes & 55.2 & 31.2 & 77.6 & 88.4 & 91.6 & 73.5 & 74.2 & 68.4 \\
    OpenVLA-OFT & Yes & 56.4 & 31.9 & 79.5 & 88.7 & 93.3 & 75.8 & 74.2 & 69.6 \\
    X-VLA~\cite{xvla} & Yes & 23.4 & \textbf{89.7} & 75.7 & 88.2 & \textbf{96.0} & 62.7 & 71.8 & 71.4 \\
    FastWAM~\cite{fastwam} & No & 16.4 & 44.5 & 68.9 & 78.2 & 53.7 & 37.7 & 60.7 & 51.5 \\
    ST-WAM~\cite{stwam} & No & 55.4 & 60.1 & 79.3 & 93.0 & 74.2 & \textbf{79.5} & \textbf{74.3} & 72.8 \\
    \midrule
    \method{} (ours) & No & \textbf{85.9} & 52.1 & \textbf{80.2} & \textbf{93.4} & 91.8 & 72.6 & 73.6 & \textbf{77.1} \\
    \bottomrule
  \end{tabular}
  \vspace{-0.4\baselineskip}
\end{table}

\subsection{Real-world OOD generalization}
\label{sec:exp-real}

\begin{table}[H]
  \vspace{-0.8\baselineskip}
  \begin{minipage}[t]{0.32\textwidth}
    \vspace{0pt}
    We compare \method{} with FastWAM on Sort and Fold under three increasingly
    challenging OOD settings (Table~\ref{tab:real}): Normal changes the
    environment, Hard additionally changes lighting and tabletop appearance,
    and Extreme also replaces task objects and containers. \method{} improves
    success in every task--difficulty combination.
  \end{minipage}\hfill
  \begin{minipage}[t]{0.66\textwidth}
    \vspace{-0.9\baselineskip}
    \centering
    \caption{Real-robot success rate (SR, \%) and mean completion time for
    successful trials (seconds).}
    \label{tab:real}
    \small
    \setlength{\tabcolsep}{3pt}
    \begin{tabular}{@{}llrrrr@{}}
      \toprule
      & & \multicolumn{2}{c}{SR (\%)} &
      \multicolumn{2}{c}{Time (s)} \\
      \cmidrule(lr){3-4}\cmidrule(lr){5-6}
      Task & Level & FastWAM & \method{} & FastWAM & \method{} \\
      \midrule
      \multirow{3}{*}{Sort}
        & Normal  & 65 & 100 & 41.1  & 42.6 \\
        & Hard    & 25 & 95  & 73.8  & 50.3 \\
        & Extreme & 10 & 75  & 79.5  & 39.8 \\
      \midrule
      \multirow{3}{*}{Fold}
        & Normal  & 60 & 100 & 114.5 & 136.3 \\
        & Hard    & 5  & 40  & 240.0 & 154.0 \\
        & Extreme & 0  & 10  & N/A   & 176.5 \\
      \bottomrule
    \end{tabular}
  \end{minipage}
  \vspace{-0.6\baselineskip}
\end{table}

On Sort, \method{}'s gains over FastWAM are 35, 70, and
65 percentage points from Normal to Extreme, yielding an average success rate
of 90.0\% versus 33.3\%. On Fold, the corresponding gains are 40, 35, and 10
points, yielding 50.0\% versus 21.7\%. Averaged across tasks at each difficulty,
\method{} reaches 100.0\%, 67.5\%, and 42.5\% success under Normal, Hard, and
Extreme, compared with 62.5\%, 15.0\%, and 5.0\% for FastWAM. Thus, the
advantage persists as visual changes are compounded, although Extreme Fold
remains challenging for both methods; FastWAM records no successful Extreme
Fold rollout, hence the N/A completion time. The largest average margin occurs
under Hard shifts, where \method{} exceeds FastWAM by 52.5 points across the
two tasks. Under Extreme shifts, it retains 42.5\% average success while
FastWAM falls to 5.0\%, suggesting that semantic and temporal evidence becomes
especially valuable as appearance cues increasingly depart from training.
Across the two tasks and three difficulty levels, \method{} raises average
success from 27.5\% to 70.0\%; over the four Hard and Extreme settings alone,
it raises the average from 10\% to 55\%.
Among successful rollouts, \method{} averages 44.2 seconds on Sort versus 64.8
seconds for FastWAM, and 145.2 versus 177.3 seconds on the Normal and Hard Fold
settings where both methods succeed. These completion times indicate that the
success-rate improvement does not require consistently longer successful
trajectories. The 39.8-second Extreme Sort result is success-conditioned:
difficult failures are excluded, whereas recoveries under milder shifts can
contribute longer completion times. Together with the Clean2Random results,
these physical experiments show that the benefit extends from controlled
simulation shifts to compounded changes in real environments, objects, and
appearance.

\ifarxivversion
\else
  \newpage
\fi
\subsection{Ablation studies}
\label{sec:exp-ablation}

\begin{wraptable}{r}{0.48\textwidth}
  \vspace{-1.2\baselineskip}
  \centering
  \caption{Ablation results on RoboTwin2.0-Full.}
  \label{tab:ablation}
  \small
  \setlength{\tabcolsep}{3pt}
  \begin{tabular*}{\linewidth}{@{\extracolsep{\fill}}lcc@{}}
    \toprule
    Variant & Clean & Randomized \\
    \midrule
    Without video expert & 88.6 & 86.2 \\
    Without history observations & 93.0 & 92.5 \\
    \method{} & \textbf{94.7} & \textbf{94.3} \\
    \bottomrule
  \end{tabular*}
  \vspace{-0.6\baselineskip}
\end{wraptable}

On RoboTwin2.0-Full, we compare \method{} with two matched ablations:
\emph{Without video expert} retains only the semantic and action experts, while
\emph{Without history observations} sets the history length to zero; all
variants use the same data and optimization budget. Removing history reduces
Clean and Randomized success by 1.7 and 1.8 points, showing the value of temporal
evidence about object displacement, interaction, and task progress. Removing
the video expert causes larger drops of 6.1 and 8.1 points, indicating that
fine-grained visual dynamics complement the semantic pathway's higher-level
state modeling.

\subsection{Component analysis}
\label{sec:exp-components}

\begin{wraptable}{r}{0.475\textwidth}
  \vspace{-1.2\baselineskip}
  \centering
  \caption{Comparison of semantic encoders on RoboTwin2.0-Full.}
  \label{tab:semantic-encoder}
  \small
  \setlength{\tabcolsep}{3pt}
  \begin{tabular*}{\linewidth}{@{\extracolsep{\fill}}lcc@{}}
    \toprule
    Semantic encoder & Clean & Randomized \\
    \midrule
    DINOv3~\cite{dinov3} & 93.3 & 92.8 \\
    V-JEPA~2.1 & \textbf{94.7} & \textbf{94.3} \\
    \bottomrule
  \end{tabular*}
  \vspace{-0.6\baselineskip}
\end{wraptable}

We first compare V-JEPA~2.1 with DINOv3 under a matched downstream token budget
(Table~\ref{tab:semantic-encoder}). Replacing
V-JEPA with DINOv3 reduces success by 1.4 points on Clean and 1.5 points on
Randomized. We match the downstream token count: V-JEPA jointly encodes eight
frames into four tubelets, whereas DINOv3 encodes four sampled frames
independently on the same spatial grid. Despite using a DINOv3 ViT-L encoder
with approximately $3.5\times$ the parameters of V-JEPA~2.1 ViT-B, the
framewise variant performs worse. This supports jointly encoded temporal
features, which expose state transitions and motion more directly than
independent image descriptors.

\begin{wraptable}{r}{0.42\textwidth}
  \vspace{-1.2\baselineskip}
  \centering
  \caption{Effect of history length on RoboTwin2.0-Full.}
  \label{tab:history-frames}
  \small
  \setlength{\tabcolsep}{3pt}
  \begin{tabular*}{\linewidth}{@{\extracolsep{\fill}}ccc@{}}
    \toprule
    History frames & Clean & Randomized \\
    \midrule
    0 & 93.0 & 92.5 \\
    4 & 93.6 & 93.0 \\
    8 & \textbf{94.7} & \textbf{94.3} \\
    \bottomrule
  \end{tabular*}
  \vspace{-0.6\baselineskip}
\end{wraptable}

We next vary only the history length (Table~\ref{tab:history-frames}). Increasing
the history from 0 to 4
frames improves Clean and Randomized success by 0.6 and 0.5 points, and
increasing it from 4 to 8 frames provides a further 1.1- and 1.3-point gain.
The larger second gain indicates that four frames do not saturate the useful
temporal context; the wider window exposes additional state changes and motion.

%% file: sec/6_conclusion.tex
\section{Conclusion}
\label{sec:conclusion}

We presented \method{}, which augments an action-only WAM with V-JEPA~2.1
representations of causal observation history and predictive semantic modeling.
\method{} achieves 94.5\% average success on RoboTwin2.0-Full, raises
Clean2Random OOD success from 10.16\% to 45.18\% over FastWAM, and improves
average success from 10\% to 55\% across the Hard and Extreme real-robot
settings. Ablations and component analyses further verify the benefits of
V-JEPA features and longer observation history, while showing that semantic and
generative video dynamics remain complementary. These results highlight
representation and temporal context as key to improving OOD generalization in
world action models.